\documentclass[11pt,a4paper]{article}
\usepackage[hyperref]{emnlp2020}
\usepackage{times}
\usepackage{latexsym}
\usepackage{multirow}
\usepackage{booktabs}
\usepackage[utf8]{inputenc}
\usepackage[T1]{fontenc}

\usepackage{xcolor}

\usepackage{microtype}

\aclfinalcopy 

\usepackage[capitalise]{cleveref}
\usepackage{todonotes}
\usepackage{booktabs,tabularx}
\def\parcite#1{\citep{#1}} 
\def\perscite#1{\citet{#1}} 

\newcommand{\wino}[0]{WinoMT}

\title{Gender Coreference and Bias Evaluation at WMT 2020}

\author{
  Tom Kocmi \thanks{\quad Part of work performed while at Charles University.}\\
  Microsoft \\
  \texttt{tomkocmi@microsoft.com} 
  \And
  Tomasz Limisiewicz \\
  Charles University in Prague \\
  Faculty of Mathematics and Physics \\
  \texttt{limisiewicz@ufal.mff.cuni.cz} \\
  \AND
  Gabriel Stanovsky \\
  The Hebrew University of Jerusalem \\
  \texttt{gabis@cse.huji.ac.il} \\
  
}

\date{}

\begin{document}
\maketitle
\begin{abstract}
Gender bias in machine translation can manifest when choosing gender inflections based on spurious gender correlations. For example, always translating doctors as men and nurses as women. This can be particularly harmful as models become more popular and deployed within commercial systems.
Our work presents the largest evidence for the phenomenon in more than 19 systems submitted to the WMT over four diverse target languages: Czech, German, Polish, and Russian.
To achieve this, we use \wino{}, a recent automatic test suite which examines gender coreference and bias when translating from English to languages with grammatical gender. We extend \wino{} to handle two new languages tested in WMT: Polish and Czech.
We find that all systems consistently use spurious correlations in the data rather than meaningful contextual information.
\end{abstract}

\section{Introduction}
Bias in machine learning occurs when systems pick up correlations which are useful for specific training \emph{datasets}, but are not indicative for the \emph{task} that the dataset represents.

In the context of machine translation (MT), gender bias can occur when translating from languages without grammatical noun genders, such as English or Turkish, to a language with gender inflections, such as  Spanish, Polish, or Czech. 
In such cases, the translation model needs to assign gender inflection in the target language based on contextual cues in the source text.
For example, when translating the English sentence 
``The \emph{doctor} asked the nurse to help \emph{her} in the operation'' to Spanish, the model 
needs to produce the female inflected ``doctora'' based on the feminine English pronoun ``her''.

Recently,~\newcite{winomt} created a challenge set and an automatic evaluation metric, dubbed~\wino, to examine whether popular MT models are capable of correctly capturing and translating such information from English into a diverse set of 8 target languages with grammatical gender.
They found that all six tested systems, composed of four commercial and two academic models, consistently relied on gender role assignments in the data regardless of context.
In our example above, models would prefer to translate the doctor using masculine inflections, despite the context suggesting otherwise.

In this work, we apply the \wino~test suite on the submissions to the News shared task of WMT 2020. In addition to testing the phenomenon on a large number of models, we extend the \wino~to the Polish and Czech languages, tackling unique language-specific challenges.
We thoroughly analyze the extent of the phenomena for the tested languages and systems, as well as its correlation with the widely-used BLEU evaluation metric~\cite{papineni2002bleu}, finding that systems with worse performance (in BLEU) make more errors for female professions than errors for male professions. On the other hand, better-performing systems (in BLEU) make more errors related to anti-stereotypical professions (e.g. female doctors, or male nurses).

Similarly to the conclusions of~\newcite{winomt}, we find that all systems consistently perform better when the source texts exhibit stereotypical gender role assignments (e.g., male doctors, female nurses) versus non-stereotypical assignments (e.g., female doctors, male nurses), indicating that these models rely on spurious correlations in their training data, rather than on more meaningful textual context.
We hope that this evaluation will be used as a standard evaluation metric for MT as a means to track the improvement of this socially important aspect of translation. 

\section{Background: \wino}
\wino~was created as a concatenation of two coreference test suites: WinoGender~\cite{winogender} and WinoBias~\cite{winobias}. 
Each instance in these datasets is a single English sentence, presenting two entities, identified by their profession (e.g., ``teacher'', ``janitor'', or ``hairdresser'') and a single pronoun referring to one of them based on the context of the sentence. For example, in the sentence ``The physician hired the secretary because \emph{he} was overwhelmed with clients'', the marked pronoun refers to the physician. In contrast, in ``The physician hired the secretary because \emph{he} had good credentials'' the pronoun likely refers to the secretary. 
Both datasets are created with an equal amount of stereotypical gender role assignments (e.g., the first example) and non-stereotypical assignments (e.g., the second example).  
Both works found that coreference systems performed much better on the stereotypical role assignments than they did on the non-stereotypical ones, concluding that systems relied on training correlations between pronoun gender and professions rather than the syntactic and semantic information in the input sentence.

\newcite{winomt} use these two corpora to test gender bias in machine translation in the following manner:
\begin{enumerate}
    \item An MT model is used to translate these corpora into a target language with grammatical gender.
    \item A language-specific, target-side morphological analysis identifies the gender of the translated entity (e.g., the physician in the first example above).
    \item The gold and predicted genders are compared between the English and target sentence. 
\end{enumerate}

Following this procedure, they tested four commercial systems and two state-of-the-art academic models on eight diverse target languages:
Spanish, French, Italian,  Hebrew, Arabic, Ukrainian, Russian, and German.
In all of their experiments, they found that similarly to coreference models, MT systems are prone to make gender-biased predictions.

In Section~\ref{sec:new-langs}, we describe our extension of \wino{} to two additional languages tested in WMT 2020: Czech and Polish, and in Section~\ref{sec:evaluation} we use the extended test suite to evaluate WMT submission on a total of four target languages: Czech, German, Polish, and Russian.

\section{New Target Languages: Polish and Czech}
\label{sec:new-langs}
In this section, we describe our extension of \wino{} for Czech and Polish. 
The methods and analyses for both target languages are done by the first two authors, who are native speakers in the respective language.

For both languages, we followed the approach of \wino{}, where translated sentences are first aligned by fast\_align~\parcite{dyer2013fastalign}, followed by automatic morphology analysis. 

Besides, we notice that the automatic alignment and existing tools sometimes fail leading to ``unknown'' gender decision. For both Czech and Polish, it could not recognise on average 10--15\% test examples. 

Fortunately, both languages have rich morphology where gender can be often identified from the word form. Therefore, we have created a list of the most often translations of each profession in all cases. Example of such a list is in \cref{tab:whitelisting_examples}. We use this list to the first check if the gender can be recognised solely based on the word form. In case that the word is not in our predefined list or if both the male and female version are possible. We revert to language-specific automatic analysis, as described below.

\begin{table*}[t]
\center
\begin{tabular}{lll}
\toprule
Profession & Gender & Possible forms \\
\midrule
Chef & Male   & šéfce, šéfka, šéfko, šéfkou, šéfku, šéfky, náčelnice, náčelnici, náčelnicí, ... \\
Chef & Female & šéf, šéfa, šéfe, šéfem, šéfovi, šéfu, náčelník, náčelníka, náčelníkem, ...  \\
Cashier & Male & pokladní, pokladního, pokladním, pokladnímu \\
Cashier & Female & pokladní \\
\bottomrule
\end{tabular}
\caption{Example of a list of possible forms for a given profession and gender (some forms are missing). Some professions have several possible translations; in this example ``chef'' has two possible translations. In Czech, most of the forms are distinct between male and female form. However, it is not always the case as can be seen for example for ``cashier'', where both male and female can have the form ``pokladní''. In those cases, we need to rely on automatic annotation based on the context of the whole sentence.}
\label{tab:whitelisting_examples}
\end{table*}

This step significantly reduced the number of unrecognised genders. In \cref{ssec:unrecognized}, we discuss the number of unrecognised genders of the profession.

\subsection{Czech analysis}

For translated professions in Czech that were not resolved by the predefined list, we use the automatic morphology tagger MorphoDiTa~\parcite{strakova14}. This tool uses a morphological dictionary and estimates regular patterns based on common form endings, by which it clusters morphological ``templates'' without linguistic knowledge of Czech. 

When analysing Czech, we ignore all examples that test neutral form, as Czech does not use neutral case as a grammatical structure allowing both genders.\footnote{In a few cases, the neutral form can be created by inaccurate translation, when replacing a profession with a place, where the professional works. For example, ``hairdresser'' can be replaced by ``hair saloon'' which is neutral in Czech.}
Additionally, we ignore a few idiosyncratic edge cases: The word ``advisee'' cannot be directly translated into Czech, while ``guest'' and ``mover'' do not have a female counterpart.

Altogether, we exclude 470 examples from \wino{}, reducing its size for Czech analysis to 3418 examples.

Lastly, certain translated professions have the same form for both male and female, for example, the word ``vedoucí'' (``supervisor'', either male or female). In such cases, our analysis cannot correctly assign correct gender. Therefore we mark these example as a correct with the use of the gold data.

\subsection{Polish analysis}
\label{sec:polish-analysis}

For translated professions in Polish that are not found in the prepared list of possible word forms, we conduct an automatic morphological tagging to find their grammatical gender. For that purpose, we use a recently released spaCy model~\cite{spacy2} with tagger for Polish~\cite{tuora-kobylinski-2019-pl-spacy}, which relies on dictionary-based morphology analysis performed by Morfeusz~\cite{wolinski-2014-morfeusz}.

Similarly to Czech, in Polish, there are no names of professions with a neutral gender. Therefore for Polish analysis, we also ignore test cases for neural form. Additionally, we do not evaluate gender for the professions that do not have a polish translation, i.e. ``advisee'' and ``mover''. This reduces the Polish testset to 3136 examples.

In Polish, it is possible to indicate gender for almost all profession names. In most cases, it can be formed by changing the suffix of the word. Nevertheless, for specific occupations, female counterparts created by derivation are rarely used and do not appear in major language dictionaries. For such professions, a feminine variant is obtained by adding a word indicating gender in front -- usually ``pani(\k{a})'' (``mrs.'') before the masculine form of the occupation name. In our evaluation, we accept both variants.

We have identified 16 professions without feminine derivations in the on-line version of the Grammatical Dictionary of Polish \cite{wolinski-kiers-2016-online}. These words are: ``appraiser'', ``driver'', ``electrician'', ``engineer'', ``firefighter'', ``investigator'', ``mechanic'', ``pathologist'', ``plumber'', ``scientist'', ``sheriff'', ``surgeon'', ``taxpayer'', ``veterinarian'', ``witness'', and ``guest''.
We decided to keep these test cases because we observed a few interesting examples of correctly translating gender for them (as discussed in section~\ref{sec:case-analysis}) and see a potential for further improvement.

\subsection{Human Annotation}

We conducted a human evaluation of gender bias for the two new languages. We sampled 300 instances from the output of all systems; each sample was annotated by two Czech and two Polish native-speakers, with a third annotator resolving differences.
Following the human evaluation protocol of \newcite{winomt}, annotators were shown an entity in English and the translated sentence. They were asked to provide the gender of the entity in the target language.

We then compared human annotations with the output of our morphological analysers in both languages. The inter-annotator agreement was high: 96.3\% for Czech and 98.6\% for Polish.
Finally, the performance of both system was good enough to support further analyses --- the Czech analyser achieved 96.3\% accuracy, while the Polish analyser achieved 98.8\%. 
Both of these numbers surpass the average performance reported in \cite{winomt} of 87\%.

Furthermore, our whitelisting approach for Czech can resolve almost all cases. From our WMT20 testsuite evaluation, it resolved all but 284 sentences out of 46,656 evaluated by all systems. We conducted a human annotation over these 284 sentences and found out that our automatic approach can correctly resolve 64.3 \% examples. 
Likewise for Polish, whitelist approach failed in only 1,533 out of 47,267 examples, where it needed to rely on the morphology evaluation. We selected 300 random sentences from this subset for human evaluation. The agreement between our morphology algorithm and annotators was 78.7\%.
We have to stress that those are the most challenging examples and most often incorrect translations.


\subsection{Unrecognized Gender}
\label{ssec:unrecognized}

When neither whitelisting nor contextual morphological analysis recognise gender properly, our automatic assigns an ``unknown'' gender. This happens mostly for erroneous translations, when the translation does not contain the profession at all (for example when ``hairdresser'' is translated as a ``hair salon''), or when an error is made by the alignment or morphology annotation.

We consider two approaches for handling such unrecognised genders. We could ignore examples with ``unknown'' gender, or we can count them as errors.
The former approach would change the ratio between male and female professions differently for each system; in other words, each system would have a testset of different size based on its performance. The latter approach, on the other hand, will punish systems for an error is in the analysis. 
We follow \wino{}, where the latter approach is selected.

To estimate the implication of this choice, in \cref{tab:unknowns} we present the average percentage of ``unknown'' genders when a gold label is male or female. The percentage is averaged across all systems.
We can notice that number of unknowns is a minimal form all languages except Russian, and the difference between unrecognised male and female professions is small. Therefore, it should not skew the results of the analysis.

We observe that the errors are usually due to the translation issue. For example, in Czech, the system with the most unrecognised genders is also the worst-performing in terms of BLEU. This system (``zlabs-nlp'') has 248 unrecognised professions out of the whole testset, while the average for all systems is 90, and it has a performance of 20.3 BLEU score, while the second-worst system has a performance of 25.3 BLEU. 

The Russian analysis cannot recognise more than 12\% of professions. We believe that this could be improved in the future with the whitelisting approach as was done for Czech and Polish.

\begin{table}[t]
\small
\center
\begin{tabular}{lcc}
\toprule
Target Language & Unknown Male & Unknown Female \\
\midrule
Czech    & 1.33\% & 1.30\%\\
German   & 1.61\% & 1.38\% \\
Polish   & 1.37\% & 1.84\% \\
Russian  & 12.89\% & 13.53\% \\
\bottomrule
\end{tabular}
\caption{The percentage of ``unknowns'' for gold male or female labels, averaged across all submissions for a target language.}
\label{tab:unknowns}
\end{table}

\section{Evaluation}
\label{sec:evaluation}

\begin{table*}[t]
\small
\center
\begin{tabular}{l|ccc|ccc|ccc|ccc}
\toprule
\multicolumn{1}{c|}{\multirow{2}{*}{\textbf{Translation System}}} &
  \multicolumn{3}{c|}{\textbf{Czech}} &
  \multicolumn{3}{c|}{\textbf{German}} &
  \multicolumn{3}{c|}{\textbf{Polish}} &
  \multicolumn{3}{c}{\textbf{Russian}} \\ \cline{2-13} 
\multicolumn{1}{c|}{} &
  \multicolumn{1}{l}{Acc} &
  \multicolumn{1}{l}{$\Delta_G$} &
  \multicolumn{1}{l|}{$\Delta_S$} &
  \multicolumn{1}{l}{Acc} &
  \multicolumn{1}{l}{$\Delta_G$} &
  \multicolumn{1}{l|}{$\Delta_S$} &
  \multicolumn{1}{l}{Acc} &
  \multicolumn{1}{l}{$\Delta_G$} &
  \multicolumn{1}{l|}{$\Delta_S$} &
  \multicolumn{1}{l}{Acc} &
  \multicolumn{1}{l}{$\Delta_G$} &
  \multicolumn{1}{l}{$\Delta_S$}  \\ \midrule
OPPO & 78.7 & 4.7 & 30.0 & \textbf{75.9} & -1.9 & 16.9 & 68.2 & 14.5 & 28.4 & 43.2 & 28.1 & 12.2 \\
zlabs-nlp & 49.9 & 38.3 & 16.3 & 71.9 & 1.1 & 8.5 & 46.1 & 50.3 & 4.3 & 36.3 & 37.8 & 6.7 \\
eTranslation & 70.9 & 11.0 & 34.5 & 71.3 & 2.9 & 18.0 & 68.8 & 11.8 & 29.0 & - & - & - \\
SRPOL & 81.2 & 3.4 & 24.3 & - & - & - & \textbf{71.2} & 12.0 & 27.6 & - & - & - \\
CUNI-Transformer & 78.0 & 5.6 & 31.8 & - & - & - & 69.8 & 14.1 & 30.6 & - & - & - \\
CUNI-DocTransformer & \textbf{83.6} & 2.2 & 22.7 & - & - & - & - & - & - & - & - & - \\
CUNI-T2T-2018 & 77.6 & 5.5 & 28.1 & - & - & - & - & - & - & - & - & - \\
UEDIN-CUNI & 72.5 & 9.4 & 28.9 & - & - & - & - & - & - & - & - & - \\
AFRL & - & - & - & 69.7 & 5.8 & 14.7 & - & - & - & - & - & - \\
Tohoku-AIP-NTT & - & - & - & 70.4 & 1.3 & 23.2 & - & - & - & - & - & - \\
UEDIN & - & - & - & 66.6 & 9.0 & 18.7 & - & - & - & - & - & - \\
WMTBiomedBaseline & - & - & - & 49.5 & 34.5 & 5.9 & - & - & - & - & - & - \\
Huoshan\_Translate & - & - & - & 63.8 & 8.0 & 24.5 & 65.7 & 18.5 & 30.7 & - & - & - \\
PROMT\_NMT & - & - & - & 65.7 & 7.2 & 17.7 & - & - & - & 44.3 & 23.8 & 14.0 \\
NICT\_Kyoto & - & - & - & - & - & - & 64.2 & 19.6 & 32.2 & - & - & - \\
SJTU-NICT & - & - & - & - & - & - & 68.2 & 15.6 & 26.1 & - & - & - \\
Tilde (1425) & - & - & - & - & - & - & 63.3 & 19.1 & 32.3 & - & - & - \\
Tilde (1430) & - & - & - & - & - & - & 64.8 & 17.7 & 23.2 & - & - & - \\
ariel197197 & - & - & - & - & - & - & - & - & - & 34.1 & 29.6 & 15.5 \\ \midrule
online-a & 63.3 & 21.7 & 21.7 & 74.5 & 0.1 & 12.5 & 53.7 & 37.8 & 21.9 & 39.1 & 35.9 & 10.2 \\ 
online-b & 56.9 & 29.7 & 19.2 & 68.3 & 2.9 & 19.4 & 57.7 & 31.9 & 21.3 & 37.8 & 36.9 & 10.4 \\
online-g & 62.0 & 22.5 & 25.9 & 62.2 & 12.0 & 16.0 & 67.3 & 17.5 & 27.7 & \textbf{47.7} & 16.2 & 17.5 \\
online-z & 72.2 & 8.2 & 30.9 & 73.6 & 0.6 & 12.4 & 65.9 & 16.0 & 35.1 & 44.4 & 25.6 & 12.5 \\
\bottomrule
\end{tabular}
\caption{The evaluation on WinoMT testset for translation systems submitted to WMT 2020. $Acc$ indicates overall gender accuracy (\% of instances the translation had the correct gender), $\Delta_G$ denotes the difference in performance (F1 score) between masculine and feminine scores, and $\Delta_S$ is the difference in accuracies between pro-stereotypical and anti-stereotypical gender role assignments. }
\label{tab:results}
\end{table*}

\begin{figure*}[h]
    \centering
    \resizebox{1.0\linewidth}{!}{\input{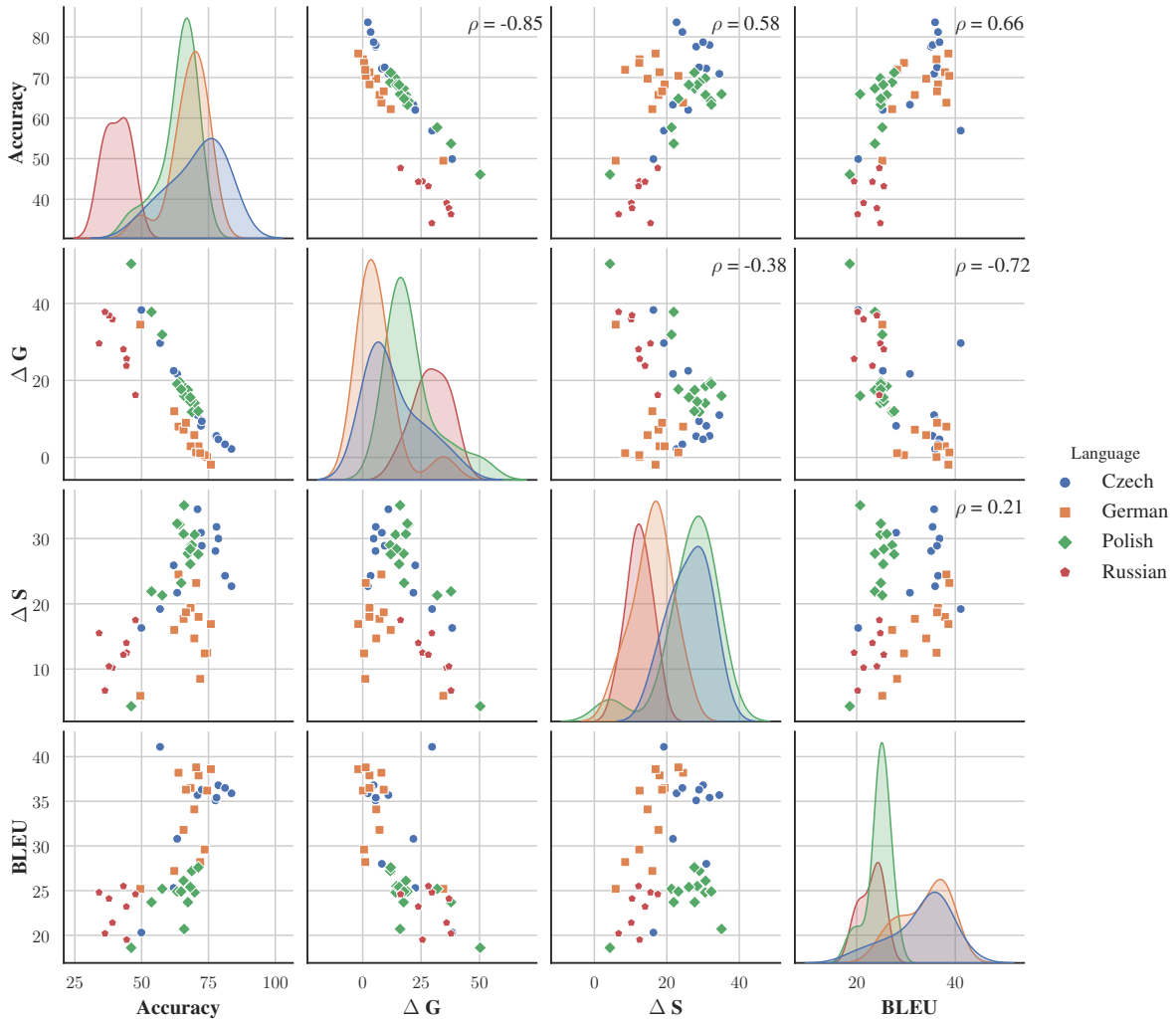}}
    \caption{Relationships between gender accuracy, $\Delta G$, $\Delta S$ and BLEU. Each point represents evaluation of one system's output. $\rho$ in the upper right is Pearson's correlation coefficient between row and column variables.}
    \label{fig:bias_vs_bleu}
\end{figure*}

\begin{table*}[h]
\centering
\begin{tabular}{p{0.29\linewidth}p{0.29\linewidth}p{0.29\linewidth}}
\toprule
\textbf{Source} & \textbf{[Target lang.] Translation} & \textbf{Phenomenon} \\ 
\midrule
The librarian waved at the \textcolor{orange}{\textbf{mechanic}} and walked away from \textcolor{red}{\textbf{her}}. & [PL] Bibliotekarka pomachała do \textcolor{blue}{\textbf{mechanika}} i od \textcolor{red}{\textbf{niej}} odeszła. & Female form for ``mechanic'' in Polish is rarely used, hence it did not occur in output of the evaluated translation systems. \\ 
\midrule
The \textcolor{orange}{\textbf{veterinarian}} informed someone that \textcolor{red}{\textbf{she}} could admit Mittens for an emergency visit. & [PL] \textcolor{red}{\textbf{Pani}} \textcolor{blue}{\textbf{weterynarz}} poinformowała właściciela, że może przyjąć Mitens na nagłą wizytę. &
Word ``weternenarz'' has a masculine gender, but translation correctly captures gender by adding word ``pani'' (``mrs.'').\\
\midrule
The \textcolor{orange}{\textbf{chief}} thanked the \textcolor{orange}{\textbf{nurse}} and handed \textcolor{blue}{\textbf{him}} a present. & [CS] \textcolor{red}{\textbf{Sestřička}} poděkovala \textcolor{blue}{\textbf{primáři}} a dala \textcolor{red}{\textbf{jí}} dárek. & There is a missmatch in translation, as female form ``jí'' refers to male ``primáři''. Moreover, it swapped actors in the translation. \\
\midrule
The \textcolor{orange}{\textbf{baker}} tried to hide the mold in the kitchen from the \textcolor{orange}{\textbf{salesperson}} but instead made \textcolor{red}{\textbf{her}} angry. & [CS] \textcolor{blue}{\textbf{Prodavač}} si všiml \textcolor{blue}{\textbf{pekařovy}} nechutné kuchyně a řekl \textcolor{blue}{\textbf{mu}}, aby přestal prodávat chleba. & In this translation, the profession is in adjective, which we are unable to detect by our analysis.\\
\bottomrule
\end{tabular}
\caption{Examples of interesting linguistic phenomena observed for translation of WinoMT sentences into new target languages. Words in \textcolor{blue}{\textbf{blue}}, \textcolor{red}{\textbf{red}}, and \textcolor{orange}{\textbf{orange}} indicate male, female and neutral entities, respectively.}
\label{tab:case-study}
\end{table*}

We continue with the analysis as described by \perscite{winomt}. For each system, we compute three metrics that represent their ability to resolve gender coreference; or how often the systems resolve the gender-based on stereotypical genders of professions. All results are in \cref{tab:results}.

\subsection{Results}

\paragraph{Overall accuracy.}
First, the overall system Accuracy (abbreviated as ``Acc'') is calculated as a percentage of instances in which the translation preserved the gender of the profession from the original English sentence. 
We find that most systems perform better than random guessing. One exception is the Russian language, where all systems perform worse than random guessing. This could be related to a problematic analysis as mentioned in \cref{ssec:unrecognized}, where all systems are penalised for ``unknown'' genders, which results in lowering their accuracy.
Overall, the system with the best accuracy is CUNI-DocTransformer on the Czech language. This system has been trained on a document-level instead of separate sentences, which may have helped it learn to resolve coreference better than sentence-level systems.
Among systems that participated in all four languages, \emph{OPPO} performs the best, also outperforming commercial systems (anonymised as ``online-X'').

\paragraph{Gender-based performance analysis.}
Second, we compute the difference $\Delta_G$ in performance (F1 score) between male and female translated professions. $\Delta_G$=0 means that the system makes an equal number of errors on both male and female professions. This should be the correct case in ideal conditions as there is an equal number of male and female examples in \wino{}. Positive $\Delta_G$ indicates that the system makes fewer errors for male professions and more errors with female professions. 
Almost all systems perform significantly better on male professions. This could be a result of training data that contains more male examples than female ones. However, we observe that many systems have $\Delta_G$ close to zero.
An interesting situation is in Czech analysis, where there is a broad range of $\Delta_G$ values.

\paragraph{Stereotypical vs non-stereotypical examples} Third, we measure the difference $\Delta_S$ in performance (F1 score) between stereotypical and non-stereotypical gender role assignments. The stereotypicality of the profession was determined based statistics provided by the US Department of Labor (see \perscite{zhao2018gender}).
$\Delta_S$ in \cref{tab:results} shows that all systems have a significantly better performance when presented with pro-stereotypical assignments (e.g., a female nurse), while their performance deteriorates when translating anti-stereotypical roles (e.g., a male receptionist). These analyses indicate that all MT systems are gender-biased, prone to translate gender inflexions based on training set correlations rather than contextual cues in specific input instances.

\subsection{Gender Bias vs BLEU}

Another interesting comparison is between the overall translation performance of a system and its observed gender bias. Unfortunately, the official WMT human annotation was not available to us at the time of writing. 
Instead, we evaluate the performance of all systems with the automatic BLEU metric~\parcite{papineni2002bleu}. The evaluation is done with official WMT20 testset \parcite{findings2020wmt} and the SacreBLEU implementation~\parcite{post2018sacrebleu}.\footnote{SacreBLEU signature is: BLEU+case.mixed+numrefs.1 +smooth.exp+tok.13a+version.1.4.14}

In \cref{fig:bias_vs_bleu} we present pairwise relationships and correlations between our metrics (Accuracy, $\Delta_G$, $\Delta_S$) and BLEU. We observe that correlation between gender accuracy and BLEU is moderately strong (Pearson's $\rho$ 0.66). There is a significant negative association between $\Delta_G$ and both gender accuracy and BLEU, meaning that systems scoring high in those metrics perform similarly well on male and female examples. We observe a low positive correlation between BLEU and $\Delta_S$ and a moderate positive correlation between gender accuracy and $\Delta_S$. It implies that the system well-performing in the automatic evaluation may overfit to examples with stereotypical gender.




\subsection{Qualitative Analysis}
\label{sec:case-analysis}

In our analysis of new target languages: Czech and Polish, we have observed several linguistic phenomena that affect gender bias of translations. We illustrate them with exemplary translations of \wino{} test sentences in \cref{tab:case-study}.

In Polish, feminine forms of some words are not commonly used, e.g., ``mechanic'' and ``veterinarian''. We provide a complete list of occupations is in \cref{sec:polish-analysis}. According to our expectations, such words are especially problematic for evaluated systems. The female gender was correctly assigned in only 1.5\% of the translations.  Interestingly, in some cases, a word indicating female gender is added, even though it is not used in a source sentence (see in \cref{tab:case-study}), such a translation is marked as correct.

\section{Conclusions}
We have extended analysis by \perscite{winomt} with Czech and Polish languages.

We showed that current systems, both commercial and academic, perform worse in gender coreference when profession in question is female. This is amplified if the system in question has lower translation quality.

Moreover, systems rely for translations on stereotypical genders of professions instead of correct gender coreference resolution. We showed that with increasing translation quality, models make fewer errors in general, but rely more often on the stereotypical genders.

This is mainly a problem with the MT training data, that usually contain more examples of stereotypical professions in contrast to anti-stereotypical ones.

\section*{Acknowledgments}

This study was supported in parts by the grants 18-24210S of the Czech Science Foundation and 825303 (Bergamot) of the European Union.
This work has been using language resources and tools stored and distributed by the LINDAT/CLARIN project of the Ministry of Education, Youth and Sports of the Czech Republic (LM2015071).

We also also would like to thank Joanna Wetesko and Maciej Biesek for discussions about the role of grammatical gender in Polish and their help with  the evaluation.

\bibliographystyle{acl_natbib}
\bibliography{emnlp2020}

\end{document}